\documentclass[10pt,twocolumn,letterpaper]{article}

\usepackage[pagenumbers]{cvpr} 

\usepackage{multirow}
\usepackage{bm}

\definecolor{cvprblue}{rgb}{0.21,0.49,0.74}
\usepackage[pagebackref,breaklinks,colorlinks,allcolors=cvprblue]{hyperref}

\def\paperID{25} 
\def\confName{CVPR}
\def\confYear{2025}

\title{Leveraging Intermediate Features of Vision Transformer for Face Anti-Spoofing}

\author{Mika Feng$^\dag$, Koichi Ito$^\dag$, Takafumi Aoki$^\dag$, Tetsushi Ohki$^\ddag$, and Masakatsu Nishigaki$^\ddag$\\
$\dag$ Graduate School of Information Sciences, Tohoku University, Japan\\
{\tt\small \{mika, ito\}@aoki.ecei.tohoku.ac.jp, aoki@ecei.tohoku.ac.jp}\\
$\ddag$ Faculty of Informatics, Shizuoka University, Japan\\
{\tt\small \{ohki, nisigaki\}@inf.shizuoka.ac.jp}
}

\begin{document}

\maketitle

\begin{abstract}
  Face recognition systems are designed to be robust against changes in head pose, illumination, and blurring during image capture.
  If a malicious person presents a face photo of the registered user, they may bypass the authentication process illegally.
  Such spoofing attacks need to be detected before face recognition.
  In this paper, we propose a spoofing attack detection method based on Vision Transformer (ViT) to detect minute differences between live and spoofed face images.
  The proposed method utilizes the intermediate features of ViT, which have a good balance between local and global features that are important for spoofing attack detection, for calculating loss in training and score in inference.
  The proposed method also introduces two data augmentation methods: face anti-spoofing data augmentation and patch-wise data augmentation, to improve the accuracy of spoofing attack detection.
  We demonstrate the effectiveness of the proposed method through experiments using the OULU-NPU and SiW datasets.
\end{abstract}

\section{Introduction}
\label{sec:intro}

Face recognition that identifies individuals based on the shape and position of their facial features is low cost since face images can be captured only by a standard camera and is highly convenient since it can perform identification without contact and any constraints \cite{Handbook-Face-Recognition}.
Since face images can easily change due to variations in head pose, illumination, and blurring during image capture, the face recognition systems are designed to be robust against such environmental changes.
Therefore, if a face photo of a registered user is presented to the face recognition system, a malicious person may bypass the authentication process illegally.
Face images of specific individual can be easily collected from the Internet, and such ``spoofing attacks'' have become a realistic threat to face recognition systems \cite{Handbook-Anti-Spoofing}. 
The major spoofing attacks are ``print attack,'' in which a printed face image is presented and ``display attack,'' in which a face video is displayed on a device such as a mobile phone and a tablet.
As mentioned above, face recognition is highly robust against environmental changes, and therefore it is necessary to detect such attacks before the authentication process to prevent spoofing attacks.


To detect spoofing attacks, it is necessary to detect minute differences between the live face image and the spoofed face image.
For example, it is necessary to extract features from the input image that can detect a global and/or local difference such as texture of paper or display device, reflection, interference fringes of display device, depth, etc.
Several methods \cite{Yu-CVPR-2020,Yu-PAMI-2020,Wang-CVPR-2022,Watanabe-APSIPA-2022,Chen-CoRR-2023,Wang-TBIOM-2022,Li-NN-2024,Zheng-IFS-2024} have been proposed to extract features inherent to spoofing attacks using Convolutional Neural Networks (CNN) and Vision Transformer (ViT) \cite{Dosovitskiy-ICLR-2021}.
All methods have high detection accuracy for simple spoofing attacks, while they have low detection accuracy for unknown spoofing attacks not included in the training data.
Since local features can be extracted in the shallow layer of ViT and global features can be extracted in the deep layer of ViT, this paper investigates spoofing attack detection that takes advantage of such characteristics of ViT.


The problem of ViT-based methods is that they are not specialized for detection of spoofing attacks.
For example, Watanabe et al.'s method \cite{Watanabe-APSIPA-2022} and Chen et al.'s method \cite{Chen-CoRR-2023} use only features extracted from the final layer of ViT and do not consider local features.
TransFAS \cite{Wang-TBIOM-2022} uses the intermediate features of ViT, however, the location of the intermediate layer from which features are extracted has not been verified.
On the other hand, this paper proposes a spoofing attack detection method that uses features extracted from the intermediate layer of ViT.
The proposed method extracts features from the intermediate layer of ViT that balance local and global features and have high generalization performance without overfitting the training data, and introduces loss functions and a score calculation method using the intermediate features.
To improve the detection accuracy against unknown spoofing attacks, we also employ Patch-wise Data Augmentation (PDA) \cite{ Watanabe-APSIPA-2022} and Face Anti-Spoofing data augmentation (FAS-Aug) \cite{Cai-IJCV-2024} for data augmentation.
We demonstrate the effectiveness of the proposed method through experiments using large-scale face image spoofing attack datasets: OULU-NPU \cite{Boulkenafet-FG-2017} and Spoofing in the Wild (SiW) \cite{Liu-CVPR-2018}.


\section{Related Work}

This section presents an overview of face spoofing attack detection, anomaly detection using intermediate features of the model, and data augmentation for face spoofing attack detection.


\subsection{Face Spoofing Attack Detection}

Major methods for detecting face spoofing attacks utilize CNN and ViT.
There are several CNN-based methods such as CDCN++ \cite{Yu-CVPR-2020}, NAS-FAS \cite{Yu-PAMI-2020}, and PatchNet \cite{Wang-CVPR-2022}.
CDCN++ \cite{Yu-CVPR-2020} extracts depth features of face images using CNN and detects spoofing attacks using these features.
NAS-FAS \cite{Yu-PAMI-2020} detects spoofing attacks using an optimal network derived from Neural Architecture Search (NAS).
PatchNet \cite{Wang-CVPR-2022} employs a patch-based approach, dividing face images into patches and extracting local features from each patch using CNN to detect spoofing attacks.
CNNs cannot always recognize the various changes that occur in a face image due to spoofing attacks.
In particular, the detection accuracy is degraded for unknown spoofing attacks that are not included in the training data.
ViT \cite{Dosovitskiy-ICLR-2021} can extract both local and global features from an image, and thus can detect local and global changes in a face image caused by spoofing attacks.
Watanabe et al. \cite{Watanabe-APSIPA-2022} introduced the data augmentation method, PDA, which is dedicated to detecting spoofing attacks, into ViT to improve the accuracy of spoofing attack detection using ViT.
Chen et al. \cite{Chen-CoRR-2023} proposed a spoofing attack detection method using Segment Anything Model (SAM) \cite{Kirillov-ICCV-2023}, which is a foundation model of image segmentation based on ViT.
TransFAS \cite{Wang-TBIOM-2022} improved the accuracy of spoofing attack detection by adding a depth feature extraction module to ViT.
Watanabe et al. \cite{Watanabe-APSIPA-2022} and Chen et al. \cite{Chen-CoRR-2023} utilize only features extracted from the deep layer of ViT to detect spoofing attacks, and thus do not consider local and global features in an optimal balance.
TransFAS \cite{Wang-TBIOM-2022} utilizes intermediate features of ViT, however, it has not been sufficiently investigated whether it necessarily uses features extracted from the suitable layer for detecting spoofing attacks.


\subsection{Anomaly Detection Using Intermediate Features}

Face spoofing attack detection can be considered as a type of anomaly detection tasks in the sense that changes contained in the image are detected.
In anomaly detection, there are approaches to improve the detection accuracy by using intermediate features, which are extracted from the intermediate layers of the model \cite{Defard-ICPR-2021,Roth-CVPR-2022}.
PaDiM \cite{Defard-ICPR-2021} utilizes intermediate features from three layers to obtain a feature set that has both local and global characteristics.
Only normal images are input to the trained model, and the features extracted from the three intermediate layers are used to estimate the distribution of normal images.
In inference, abnormality is detected based on the distance between the distribution of normal images and the features of the input image extracted from the three intermediate layers.
PatchCore \cite{Roth-CVPR-2022} extracts features from the intermediate layers of the trained model to obtain local features that do not overfit the training data.
In general, the features output from the final layer of a trained model are used for anomaly detection using the distance between the features of the input image and those of normal images.
The features in the final layer may lack local characteristics that are important for anomaly detection, and may contain features that are biased towards the pre-trained task.
PatchCore \cite{Roth-CVPR-2022} addresses this problem by using features extracted from the intermediate layers of the trained model.
Normal images are input to the trained model, and the features extracted from the intermediate layers are stored in a memory bank.
In inference, the image is input to the trained model and the degree of anomaly is calculated based on the distance between the features extracted from the intermediate layer and the features in the memory bank.
As described above, the approach of using intermediate features in anomaly detection is also effective in spoofing attack detection.
In particular, to detect unknown spoofing attacks, it is necessary to obtain features with high generalization performance that do not overfit the training data, and therefore the use of features extracted in the intermediate layers of ViT is considered effective.


\begin{figure*}[t]
  \centering
  \includegraphics[width=\linewidth]{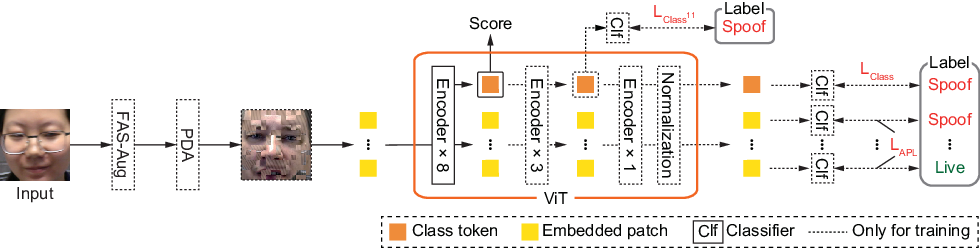}
  \caption{Overview of the proposed method: Network architecture and loss functions.}
  \label{fig:proposed}
\end{figure*}

\subsection{Data Augmentation for Face Spoofing Attack Detection}

Although detection accuracy has been improved through model refinement, there is a problem that detection accuracy is low for spoofing attacks that are not included in the training data.
Therefore, data augmentation that adds variations to the training data has been considered to improve the accuracy to address unknown spoofing attacks \cite{Watanabe-APSIPA-2022,Cai-IJCV-2024}.
Watanabe et al. \cite{Watanabe-APSIPA-2022} proposed Patch-wise Data Augmentation (PDA), which improves the generalization performance of a model by making the training data more difficult to discriminate.
PDA improves detection accuracy by replacing parts of the spoofed attack image with patches of the live image, and then learning to recognize the replaced image as ``Spoof.''
Cai et al. \cite{Cai-IJCV-2024} proposed Face Anti-Spoofing data augmentation (FAS-Aug), which is a data augmentation specialized for spoofing attack detection.
FAS-Aug improves detection accuracy by adding characteristics that appear during the image acquisition process, such as color change, noise from printing, and interference fringes, to the training data.
Since these data augmentations are effective in detecting spoofing attacks, we consider that the combined implementation of PDA and FAS-Aug can improve detection accuracy.



\section{Proposed Method}

The proposed method consists of four key components: (i) the use of features extracted from the intermediate layer of ViT \cite{Dosovitskiy-ICLR-2021} (Sect. \ref{sec:vit}), (ii) data augmentation to deal with unknown spoofing attacks (sect. \ref{sec:da}), (iii) design of loss functions specialized for spoofing attacks (Sect. \ref{sec:loss}), and (iv) score calculation based on intermediate features (Sect. \ref{sec:score}), into a face spoofing attack detection based on ViT in order to improve the detection accuracy against unknown spoofing attacks.
First, the data augmentation methods: FAS-Aug \cite{Cai-IJCV-2024} and PDA \cite{Watanabe-APSIPA-2022}, are applied to the input image.
Note that FAS-Aug and PDA are applied only in training, not in inference.
Next, the image is divided into patches and input to ViT for feature extraction.
ViT used in the proposed method consists of 12 layers of encoder blocks and a normalization layer in the final layer.
For training, we use the loss to the class token in the 11th layer, the loss to the class token in the final layer, and Attention-Weighted Patch Loss (APL) \cite{Watanabe-APSIPA-2022}.
For inference, the class token output from the encoder block in the 8th layer is used to compute a score for detecting spoofing attacks.
We describe each component of the proposed method in the following.


\subsection{Intermediate Features from ViT}
\label{sec:vit}

Fig. \ref{fig:proposed} shows the network architecture based on ViT used in the proposed method.
The proposed method uses ViT (ViT-Base) \cite{Dosovitskiy-ICLR-2021}, in which an input image of $224 \times 224$ pixels is divided into $16 \times 16$ patches and input into an encoder block consisting of 12 layers.
We employ parameters pre-trained on a large-scale and non-public Google dataset \cite{Dosovitskiy-ICLR-2021} as initial values, and perform fine tuning on the training data of a face spoofing attack detection dataset.
In original ViT, image classification is performed using class tokens output at the final layer.
The proposed method outputs not only the class token but also the features of each patch in the final layer to detect signatures of spoofing attacks that appear in local regions.
The proposed method also employs the class token extracted in the 8th encoder block to calculate a score for spoofing attack detection.
This is because intermediate features that balance global and local features of an image are effective in anomaly detection \cite{Defard-ICPR-2021,Roth-CVPR-2022}.
The effectiveness of the 8th encoder block for detecting spoofing attacks has been empirically confirmed, and details can be found in Sect. \ref{sec:ablation_feature}.
In addition to calculating scores using intermediate features, the proposed method introduces loss for class tokens extracted from the 11th encoder block.
We have empirically confirmed that the detection accuracy is improved by calculating the loss in the 11th encoder block as well, instead of calculating the loss only in the final layer, and details can be found in Sect. \ref{sec:ablation_loss}.


\subsection{Data Augmentation}
\label{sec:da}

\begin{figure}[t]
  \centering
  \includegraphics[width=.95\linewidth]{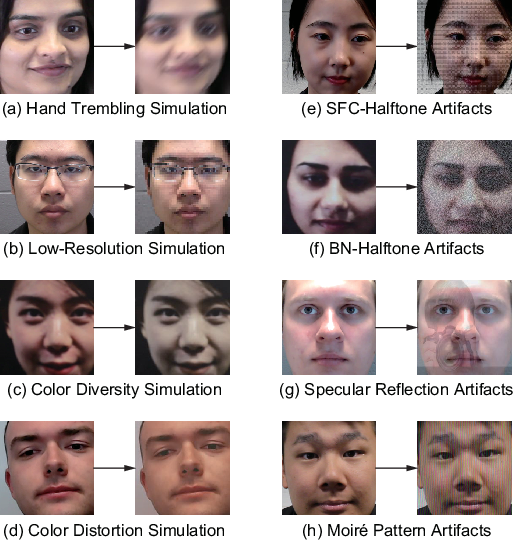}
  \caption{Example of face images after applying FAS-Aug.}
  \label{fig:FAS-Aug}
\end{figure}

\begin{figure}[t]
  \centering
  \includegraphics[width=.85\linewidth]{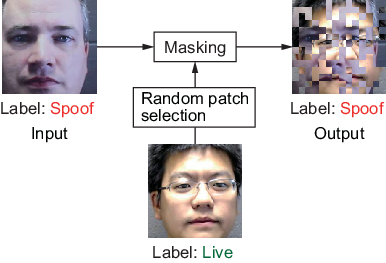}
  \caption{Overview of the procedure of Live Patch Mask in PDA.}
  \label{fig:PDA}
\end{figure}

The proposed method employs a combination of FAS-Aug \cite{Cai-IJCV-2024} and PDA \cite{Watanabe-APSIPA-2022}, which are data augmentation methods to improve the accuracy of spoofing attack detection.
FAS-Aug \cite{Cai-IJCV-2024} consists of 8 types of data augmentation that emulate photography noise, print attack, and display attack.
As photography noise, (a) hand trembling simulation, (b) low-resolution simulation, and (c) color diversity simulation are added to the image.
Since these noises occur in both live and spoofed images, the labels are not replaced when these noises are added.
As print-attack noise, (d) color distortion simulation, (e) SFC-halftone artifacts, and (f) BN-halftone artifacts are added to the image.
When these are added, the label is replaced by ``Spoof.''
As display-attack noise, (g) specular reflection artifacts and (h) moir\'{e} pattern artifacts are added to the image.
When these are added, the label is replaced by ``Spoof.''
The parameters used in FAS-Aug are in accordance with \cite{Cai-IJCV-2024}.
The proposed method randomly selects one of the 8 data augmentations and applies it to the input image.
Fig. \ref{fig:FAS-Aug} shows the example of images to which the 8 data augmentations included in FAS-Aug are applied.
PDA \cite{Watanabe-APSIPA-2022} is a data augmentation method applied to each patch of ViT, where the proposed method employs only ``Live Patch Mask'' in PDA \cite{Watanabe-APSIPA-2022} because of its effectiveness in detecting unknown spoofing attacks.
As shown in Fig. \ref{fig:PDA}, some patches of the spoofed image is replaced by patches of the live image, and the model is trained to recognize the patch of the spoofed image as ``Spoof,'' the replaced patch of the live image as ``Live,'' and the entire image as ``Spoof.''
This mixing of a spoofed attack image with a live image makes detection more difficult, thereby enhancing the learning of the model.
We replace each patch with a probability of 0.5 in the proposed method.



\subsection{Loss Function}
\label{sec:loss}

In training of the network used in the proposed method as shown in Fig. \ref{fig:proposed}, we employ the following 2-class classification loss functions: $\mathcal{L}_{Class}$ for the class token output from the final layer of ViT, $\mathcal{L}_{APL}$ for each patch weighted by the attention map of the class token output from the 12th encoder block of ViT \cite{Watanabe-APSIPA-2022}, and $\mathcal{L }_{Class^{11}}$ for the class token output from the 11th encoder block of ViT.
$\mathcal{L}_{Class}$ is the loss function used in standard ViT-based classification.
$\mathcal{L}_{APL}$ was proposed in \cite{Watanabe-APSIPA-2022} to take into account patch-wise spoofing attack detection.
For details, refer to \cite{Watanabe-APSIPA-2022}.
As mentioned above, we have empirically confirmed that $\mathcal{L}_{Class^{11}}$ refines the features of the class token output from the 8th encoder block used to calculate the score.
For details of experimental verification, refer to Sect. \ref{sec:ablation_loss}.
For each 2-class classification loss, we use L2-constrained softmax loss \cite{Ranjan-CoRR-2017} instead of Softmax Loss.
By constraining the L2 norm of the feature vectors to be $\alpha$, the feature vectors are trained equally without bias toward either ``Live'' or ``Spoof.''
In L2-constrained softmax loss, we minimize the loss calculated by 
\begin{equation}
  {\rm L2Softmax} = -\frac{1}{N} \sum_{i=1}^N \log \frac{e^{W_{y_i}^{\top} f(\bm{x}_i)+b_{y_i}}}{\sum_{j=1}^C e^{W_j^{\top} f(\bm{x}_i)+b_j}},
\end{equation}
\begin{equation*}
  {\rm s.t.} \ \ \| f(\bm{x}_i) \|_2 = \alpha, \quad \forall i = 1, 2, \ldots, N,
\end{equation*}
where $\bm{x}_i$ is the input image, $N$ is the size of the mini-batch, $W$ is the weight matrix of fully-connected layers, $f(\bm{x}_i)$ is the feature vector, $C$ is the number of classes, $\alpha$ is a hyperparameter, $W_{y_i}$ is the column vector corresponding to the correct label $y_i$ in $W$, and $b_{y_i}$ is the bias corresponding to $y_i$.
The total loss function $\mathcal{L}_{Overall}$ is given by
\begin{equation}
  \mathcal{L}_{Overall} = \mathcal{L}_{Class}+\mathcal{L}_{APL}+\mathcal{L}_{Class^{11}},
\end{equation}
where each loss function, $\mathcal{L}_{Class}$, $\mathcal{L}_{APL}$, and $\mathcal{L}_{Class^{11}}$, is calculated by L2-constrained softmax loss.


\subsection{Score Calculation}
\label{sec:score}

The detection of spoofing attacks is based on the cosine similarity between the features of the class tokens output from the 8th encoder block of ViT, rather than the final layer of ViT.
The ``Live'' images of the training data are input to ViT after training, and the class token output from the 8th encoder block of ViT is obtained.
These class tokens are used as the reference vectors of the ``Live'' image to be compared when calculating the cosine similarity.
In inference, a face image is input to ViT, and the class token output from the 8th encoder block is obtained.
$Score$ is the maximum value of the cosine similarity between the class tokens of the input image and all the reference vectors obtained from the ``Live'' images in the training data.
If $Score$ is greater than or equal to a threshold, the image is considered ``Live,'' otherwise, it is considered ``Spoof.''
We have empirically confirmed that the class token output from the 8th encoder block is more accurate in detecting spoofing attacks than the class tokens output from the other layers of ViT, including the final layer.
For details, refer to Sect. \ref{sec:ablation_feature}.


\section{Experiments and Discussion}

This section describes experiments to demonstrate the effectiveness of the proposed method in detecting face spoofing attacks.


\subsection{Dataset}

We use the OULU-NPU \cite{Boulkenafet-FG-2017} and SiW \cite{Liu-CVPR-2018} datasets in the following experiments.
OULU-NPU consists of 4,950 videos taken from 55 subjects.
Each frame is $1,080 \times 1,920$ pixels and each video is captured at 30 fps for about 15 seconds.
The ``Live'' videos are captured using the front cameras of 6 mobile devices, i.e., Samsung Galaxy S6 edge, HTC Desire EYE, MEIZU X5, ASUS Zenfone Selfie, Sony XPERIA C5 Ultra Dual, and OPPO N3, in 3 sessions with different illumination conditions and background scenes
The print attack uses a paper printed by two types of printers, and the display attack uses two types of display devices.
SiW consists of 4,478 videos taken from 165 subjects.
For each subject, there are 8 real videos and up to 20 spoofed videos.
Each frame is $1,920 \times 1,080$ pixels and each video is captured at 30 fps for about 15 seconds.
The ``Live'' videos are captured under various lighting conditions while the subjects move their heads back and forward, rotate their heads, and change their facial expressions.
The print attack uses two types of paper, and the display attack uses four types of display devices.


\subsection{Evaluation Protocol}

OULU-NPU \cite{Boulkenafet-FG-2017} and SiW \cite{Liu-CVPR-2018} provide evaluation protocols to evaluate the generalization capabilities of spoofing attack detection methods.
This paper also conducts experiments according to the evaluation protocols provided by each dataset.
The following provides an overview of the evaluation protocols for each dataset.


OULU-NPU \cite{Boulkenafet-FG-2017} provides 4 evaluation protocols.
Protocol 1 evaluates generalization performance for illumination and background changes.
2 out of 3 sessions are used for training, and the remaining is used for evaluation.
Protocol 2 evaluates generalization performance for the types of paper and display devices.
One of each of the 2 types of paper and display devices is used for training, and the remaining is used for evaluation.
Protocol 3 evaluates generalization performance for different mobile cameras.
5 out of 6 mobile cameras are used for training and the remaining is used for evaluation.
The detection accuracy is calculated for 6 different patterns with different input sensors used during the evaluation, and evaluated by their mean and standard deviation.
Protocol 4 evaluates the generalization performance for all variations of Protocol 1 to 3.
The detection accuracy is calculated for 6 patterns with different sessions and mobile cameras, and their average and standard deviation are used for evaluation.


SiW \cite{Liu-CVPR-2018} provides 3 evaluation protocols.
Protocol 1 evaluates generalization performance for changes in pose and facial expression.
The changes in pose and facial expression are small for the first 60 frames, while the changes are large for the subsequent frames.
The first 60 frames in the training data are used for training, and all frames in the test data are used for evaluation.
Protocol 2 evaluates generalization performance for the types of display devices used in display attacks.
3 of the 4 types of display devices are used for training, and the remaining is used for evaluation.
The detection accuracy is calculated for four patterns using different display devices, and their average and standard deviation are used for evaluation.
Protocol 3 evaluates the generalization performance against unknown spoofing attacks.
Either the print attack or the display attack is used for training, and the other is used for evaluation.
The detection accuracy is calculated for two patterns with different types of spoofing attacks, and their average and standard deviation are used for evaluation.


\subsection{Experimental Condition}

We describe the experimental conditions in this paper.
Since videos in OULU-NPU \cite{Boulkenafet-FG-2017} and SiW \cite{Liu-CVPR-2018} contain objects that can easily be recognized as spoofing attacks, face regions are extracted from each frame of the videos and used as input images.
For OULU-NPU \cite{Boulkenafet-FG-2017}, face regions are extracted from each frame using MTCNN \cite{Xiang-ICISCE-2017} and resized to $224 \times 224$ pixels.
For SiW \cite{Liu-CVPR-2018}, the bounding box for face regions is provided, and the face regions are extracted based on the bounding box and resized to $224 \times 224$ pixels.
Then, the pixel values of the input images are normalized with zero mean and unit variance for each channel.
The proposed method assumes that the input is a single image, not a video.
Therefore, 10 frames are randomly extracted from the training or evaluation videos for each evaluation protocol and used as input images for training or evaluation in the experiments as in \cite{Boulkenafet-FG-2017}.


In training, we apply FAS-Aug \cite{Cai-IJCV-2024} with a probability of ${P}_{FAS-Aug} = 0.2$, randome flliping with a probability of 0.5, the brightness changes on a random scale, and PDA \cite{Watanabe-APSIPA-2022} with probability ${P}_{PDA} = 0.2$ to the input image, where these probabilities are empirically determined.
Note that FAS-Aug and PDA are not applied in epochs after the loss drops below 0.001 to avoid overfitting on images to which data augmentation has been applied.
The image obtained from the above process is divided into patches and input to ViT of the proposed method.
As described in Sect. \ref{sec:vit}, the proposed method uses the pre-trained model of ViT-Base \cite{Dosovitskiy-ICLR-2021} and performs fine tuning on the OULU-NPU or SiW training data.
The size of the mini-batch is 8, the learning rate is 0.0001, the number of epochs is 200, and Nesterov Accelerated Gradient \cite{Nesterov-USSR-1983} is used as optimizer.


We use $Score$ to detect spoofing attacks, as described in Sect. \ref{sec:score}.
The input image is considered ``Live'' if $Score$ is greater than or equal to a threshold, otherwise it is considered ``Spoof.''
The threshold is set at the score where the False Acceptance Rate (FAR) and False Rejection Rate (FRR) intersect, following \cite{Watanabe-APSIPA-2022}.


In this paper, we conduct 5 experiments to demonstrate the effectiveness of the proposed method.
First, we conduct the ablation study on (i) intermediate features of ViT used for score calculation (Sect. \ref{sec:ablation_feature}), (ii) intermediate features of ViT used for loss calculation (Sect. \ref{sec:ablation_loss}), (iii) loss functions (Sect. \ref{sec:ablation_lossfunction}), and (iv) data augmentation (Sect. \ref{sec:ablation_da}).
In the above experiments, we employ OULU-NPU protocol 4 and SiW protocol 3.
Next, to evaluate the effectiveness of the proposed method in detecting face spoofing attacks, we conduct (v) comparison between the proposed method and conventional methods using OULU-NPU and SiW (Sect. \ref{sec:comparison}).


\subsection{Evaluation Metrics}

In this experiment, we use Attack Presentation Classification Error Rate (APCER), Bona Presentation Classification Error Rate (BPCER), and Average Classification Error Rate (ACER).
APCER measures the maximum false acceptance rate for each spoofing attack and is calculated by
\begin{equation}
  \mathrm{APCER}=\max_{PA}\left(1-\frac{1}{N_{PA}} \sum_{i=1}^{N_{PA}} R_i\right),
\end{equation}
where $PA$ is a set of spoofing attack types, $N_{PA}$ is the number of spoofed images, and $R_i$ is a function that takes 1 if the attack is classified as ``Spoof'' and 0 if it is classified as ``Live.''
BPCER is the false rejection rate for live images and is calculated by
\begin{equation}
  \mathrm{BPCER}=\frac{1}{N_{BF}} \sum_{i=1}^{N_{BF}} R_i,
\end{equation}
where $N_{BF}$ is the number of live images.
ACER is the average of APCER and BPCER and is calculated by
\begin{equation}
  \mathrm{ACER}=\frac{\mathrm{APCER}+\mathrm{BPCER}}{2}.
\end{equation}
The smaller the value of these evaluation metrics, the higher the detection accuracy of spoofing attacks.


\subsection{Exp. (i): Intermediate Features of ViT Used for Score Calculation}
\label{sec:ablation_feature}

In this experiment, we evaluate the effectiveness of score calculation using the class token output from the 8th encoder block.
We do not introduce $\mathcal{L}_{Class^{11}}$ using the class token output from the 11th encoder block and data augmentation of FAS-Aug and PDA to evaluate only the effectiveness of the score calculation.
We compare ACER when the class token output from the 6th to 12th encoder blocks and the final normalization layer is used to calculate the score. 
Note that the class token output by the 1st to 5th encoder blocks is not used in this experiment, since they do not exhibit sufficient classification performance.
To verify the improvement in detection accuracy for various spoofing attacks, protocol 4 of OULU-NPU and protocol 3 of SiW are used.
Table \ref{tbl:Ablation_1} shows the experimental results, where the highest accuracy is shown in bold, and the second highest accuracy is underlined.
When the class token output from the 8th encoder block is used to calculate the score, the lowest ACER is obtained for both datasets.
This is because the features output from the 8th encoder block do not overfit the training data, and they have a good balance of local and global features suitable for detecting spoofing attacks.
In the following experiments, the proposed method uses the class token output from the 8th layer encoder block to compute the score.


\begin{table}[t]
  \centering
  \caption{Ablation study when changing the features used for score calculation, where the values indicate ACER [\%]$\downarrow$.}
  \label{tbl:Ablation_1}
  \scalebox{0.8}[0.8]{
  \begin{tabular}{ccc}
    \hline
    Features & OULU-NPU Prot. 4 & SiW Prot. 3 \\
    \hline
    Encoder Block \#6 & \underline{9.75$\pm$6.24} & 4.1$\pm$1.04 \\
    Encoder Block \#7 & 10.42$\pm$6.95 & \underline{3.07$\pm$0.08} \\
    Encoder Block \#8 & {\bf 8.83$\pm$7.51} & {\bf 2.2$\pm$0.57} \\
    Encoder Block \#9 & 12.46$\pm$8.02 & 3.45$\pm$1.7 \\
    Encoder Block \#10 & 13.13$\pm$8.95 & 3.13$\pm$2.08 \\
    Encoder Block \#11 & 13.46$\pm$8.83 & 3.37$\pm$2.43 \\
    Encoder Block \#12 & 15.17$\pm$8.37  & 3.58$\pm$2.81 \\
    Normalization Layer & 15.25$\pm$8.03  & 3.6$\pm$2.91 \\
    \hline
  \end{tabular}
  }
\end{table}

\subsection{Exp. (ii): Intermediate Features of ViT Used for Loss Calculation}
\label{sec:ablation_loss}

In this experiment, we evaluate the effectiveness of loss calculation using the class token output from the 11th encoder block, i.e., $\mathcal{L}_{Class^{11}}$.
In this experiment, we use the class token output from the 8th to 12th encoder blocks since the class token output from the 8th encoder block is used to calculate the score.
In addition, we do not introduce the data augmentation of FAS-Aug and PDA in this experiment since we only evaluate the effectiveness of the 2-class classification loss.
As in Exp. (i), we employ OULU-NPU protocol 4 and SiW protocol 3 in this experiment.
Table \ref{tbl:Ablation_2} shows the experimental results, where the highest accuracy is shown in bold, and the second highest accuracy is underlined.
The lowest ACER is obtained when a 2-class classification loss is introduced in the class token output from the 11th encoder block.
This result indicates that the class tokens output from the 8th encoder block can be refined by adding constraints not only to the class tokens output from the final layer, but also to the class tokens output from the 11th encoder block.



\begin{table}[t]
  \centering
  \caption{Ablation study when changing the features used for loss calculation, where the values indicate ACER [\%]$\downarrow$.}
  \label{tbl:Ablation_2}
  \scalebox{0.8}[0.8]{
  \begin{tabular}{ccc}
    \hline
    Features & OULU-NPU Prot. 4 & SiW Prot. 3 \\
    \hline
    Encoder Block \#8 & 10.46$\pm$5.60 & 3.02$\pm$2.37 \\
    Encoder Block \#9 & 6.54$\pm$5.09 & 3.07$\pm$1.74 \\
    Encoder Block \#10 & 6.13$\pm$3.81 & 2.63$\pm$0.73 \\
    Encoder Block \#11 & {\bf 5.54$\pm$3.39} & {\bf 2.41$\pm$0.89} \\
    Encoder Block \#12 & \underline{5.71$\pm$4.83} & \underline{2.59$\pm$0.08} \\
    \hline
  \end{tabular}
  }
\end{table}

\subsection{Exp. (iii): Loss Functions}
\label{sec:ablation_lossfunction}

In this experiment, we evaluate the effectiveness of the loss functions: $\mathcal{L}_{APL}$ and $\mathcal{L}_{Class^{11}}$, introduced in the proposed method in addition to the basic loss function $\mathcal{L}_{Class}$ of ViT.
As in Exp. (i) and (ii), we do not introduce FAS-Aug and PDA, and employ OULU-NPU protocol 4 and SiW protocol 3.
Table \ref{tbl:Ablation_3} shows the experimental results, where the highest accuracy is shown in bold, and the second highest accuracy is underlined.
OULU-NPU protocol 4 has the lowest ACER when $\mathcal{L}_{Class^{11}}$ and $\mathcal{L}_{APL}$ are introduced.
SiW protocol 3 has the lowest ACER when $\mathcal{L}_{Class^{11}}$ is introduced, while there is no significant difference depending on the loss functions compared to OULU-NPU protocol 4.
Since the introduction of $\mathcal{L}_{Class^{11}}$ and $\mathcal{L}_{APL}$ has significantly improved ACER for OULU-NPU protocol 4, the proposed method employs $\mathcal{L}_{Class^{11}}$ and $\mathcal{L}_{APL}$ in the following experiments.

\begin{table}[t]
  \centering
  \caption{Ablation study on loss functions, where the values indicate ACER [\%]$\downarrow$.}
  \label{tbl:Ablation_3}
  \scalebox{0.8}[0.8]{
  \begin{tabular}{ccccc}
    \hline
    $\mathcal{L}_{Class}$ & $\mathcal{L}_{Class^{11}}$ & $\mathcal{L}_{APL}$ & OULU-NPU Prot. 4 & SiW Prot. 3 \\
    \hline
    \checkmark & & & 9.46$\pm$8.01 & \underline{1.73$\pm$0.33} \\
    \checkmark & & \checkmark & \underline{8.83$\pm$7.51} & 2.2$\pm$0.57 \\
    \checkmark & \checkmark & & 9.0$\pm$6.0 & {\bf 1.63$\pm$0.19} \\
    \checkmark & \checkmark & \checkmark & {\bf 5.54$\pm$3.39} & 2.41$\pm$0.89 \\
    \hline
  \end{tabular}
  }
\end{table}

\subsection{Exp. (iv): Data Augmentation}
\label{sec:ablation_da}

In this experiment, we evaluate the effectiveness of the data augmentation methods: FAS-Aug \cite{Cai-IJCV-2024} and PDA \cite{Watanabe-APSIPA-2022}, introduced in the proposed method.
To verify the potential of data augmentation methods in detecting spoofing attacks, this experiment compares the detection accuracy of ViT with and without the intermediate features.
ViT without the intermediate features is a method that calculates the score using the class token output from the final normalization layer and uses $\mathcal{L}_{Class}$ and $\mathcal{L}_{APL}$ as the loss function.
ViT with intermediate features, \ie, the proposed method, is a method that calculates the score using the class token output from the 8th encoder block and uses $\mathcal{L}_{Class}$, $\mathcal{L}_{Class^{11}}$, and $\mathcal{L}_{APL}$ as the loss function. 
As in other ablation studies, we employ OULU-NPU protocol 4 and SiW protocol 3.
Table \ref{tbl:Ablation_4} shows the experimental results, where the highest accuracy is shown in bold, and the second highest accuracy is underlined.
For ViT without the intermediate features, ACER is the lowest for both protocols when FAS-Aug and PDA are introduced.
In particular, OULU-NPU protocol 4 exhibits significantly low ACER by introducing data augmentation methods.
For ViT with intermediate features, OULU-NPU protocol 4 exhibits the lowest ACER when FAS-Aug is introduced, and SiW protocol 3 exhibits the lowest ACER when FAS-Aug and PDA are introduced.
These results indicate that FAS-Aug which adds spoofing attack features not included in the training data and PDA which makes attack detection more difficult are effective since FAS-Aug and PDA improve ACER regardless of whether intermediate features are used or not.
The effectiveness of ViT with the intermediate features proposed in this paper has also been demonstrated, since the combination of intermediate features of ViT and data augmentation methods accelerates the improvement of ACER.


\begin{table}[t]
  \centering
  \caption{Ablation study on data augmentation, where ``ViT'' indicates with and without using the intermediate features, and the values indicate ACER [\%]$\downarrow$.}
  \label{tbl:Ablation_4}
  \scalebox{0.8}[0.8]{
  \begin{tabular}{ccccc}
    \hline
    ViT & FAS-Aug & PDA & OULU-NPU Prot. 4 & SiW Prot. 3 \\
    \hline
    & & & 15.25$\pm$8.03 & 3.6$\pm$2.91 \\
    & & \checkmark & 10.42$\pm$6.76 & 3.08$\pm$2.57 \\
    & \checkmark & & 7.63$\pm$5.19 & 4.08$\pm$3.49 \\
    & \checkmark & \checkmark & 6.08$\pm$2.76 & 2.07$\pm$1.63 \\ 
    \checkmark & & & 5.54$\pm$3.39 & 2.41$\pm$0.89 \\
    \checkmark & & \checkmark & 5.5$\pm$4.15 & 2.0$\pm$0.45 \\
    \checkmark & \checkmark & & {\bf 2.21$\pm$2.6} & \underline{1.34$\pm$0.58} \\
    \checkmark & \checkmark & \checkmark & \underline{2.54$\pm$2.38} & {\bf 0.83$\pm$0.13} \\
    \hline
  \end{tabular}
  }
\end{table}
\begin{table}[t]
  \centering
  \caption{Experimental results of each method for OULU-NPU (Unit: \%).}
  \label{tbl:results_OULU}
  \scalebox{0.8}[0.8]{
    \begin{tabular}{ccccc}
    \hline
    Prot. & Method & APCER$\downarrow$ & BPCER$\downarrow$ & ACER$\downarrow$ \\
    \hline
    \multirow{9}{*}{1} 
     & CDCN++ \cite{Yu-CVPR-2020} & \underline{0.4} & {\bf 0.0} & 0.2 \\ 
     & NAS-FAS \cite{Yu-PAMI-2020} & \underline{0.4} & {\bf 0.0} & 0.2 \\ 
     & PatchNet \cite{Wang-CVPR-2022} & {\bf 0.0} & {\bf 0.0} & {\bf 0.0} \\ 
     & Watanabe \cite{Watanabe-APSIPA-2022} & 3.7 & 2.4 & 3.0 \\ 
     & Chen \cite{Chen-CoRR-2023} & --- & --- & \underline{0.1} \\
     & TransFAS \cite{Wang-TBIOM-2022} & 0.8 & {\bf 0.0} & 0.4 \\
     & Li \cite{Li-NN-2024} & \underline{0.4} & {\bf 0.0} & 0.2 \\
     & Proposed w/o PDA & 1.21 & 1.08 & 1.15 \\
     & Proposed w/ PDA & 0.92 & \underline{0.5} & 0.71 \\ 
     \hline
    \multirow{9}{*}{2} 
     & CDCN++ \cite{Yu-CVPR-2020} & 1.8 & 0.8 & 1.3 \\ 
     & NAS-FAS \cite{Yu-PAMI-2020} & 1.5 & 0.8 & 1.2 \\ 
     & PatchNet \cite{Wang-CVPR-2022} & \underline{1.1} & 1.2 & 1.2 \\ 
     & Watanabe \cite{Watanabe-APSIPA-2022} & \underline{1.1} & 0.7 & \underline{0.9} \\ 
     & Chen \cite{Chen-CoRR-2023} & --- & --- & 1.1 \\
     & TransFAS \cite{Wang-TBIOM-2022} & 1.5 & \underline{0.5} & 1.0 \\
     & Li \cite{Li-NN-2024} & 1.5 & \underline{0.5} & 1.0 \\
     & Proposed w/o PDA & 1.36 & 0.69 & 1.02 \\
     & Proposed w/ PDA & {\bf 0.69} & {\bf 0.39} & {\bf 0.54} \\
     \hline
    \multirow{9}{*}{3} 
     & CDCN++ \cite{Yu-CVPR-2020} & 1.7$\pm$1.5 & 2.0$\pm$1.2 & 1.8$\pm$0.7 \\ 
     & NAS-FAS \cite{Yu-PAMI-2020} & 2.1$\pm$1.3 & 1.4$\pm$1.1 & 1.7$\pm$0.6 \\ 
     & PatchNet \cite{Wang-CVPR-2022} & 1.8$\pm$1.47 & \underline{0.56$\pm$1.24} & 1.18$\pm$1.26 \\ 
     & Watanabe \cite{Watanabe-APSIPA-2022} & 1.2$\pm$1.0 & 0.7$\pm$1.0 & 1.0$\pm$1.0 \\ 
     & Chen \cite{Chen-CoRR-2023} & --- & --- & 1.4$\pm$1.21 \\
     & TransFAS \cite{Wang-TBIOM-2022} & {\bf 0.6$\pm$0.7} & 1.1$\pm$2.5 & \underline{0.9$\pm$1.1} \\
     & Li \cite{Li-NN-2024} & \underline{0.7$\pm$0.9} & 1.1$\pm$2.7 & 0.9$\pm$1.3 \\
     & Proposed w/o PDA & 1.39$\pm$0.67 & 0.86$\pm$0.52 & 1.13$\pm$0.57 \\
     & Proposed w/ PDA & 0.85$\pm$0.62 & {\bf 0.33$\pm$0.3} & {\bf 0.59$\pm$0.45} \\
     \hline
     \multirow{9}{*}{4} 
     & CDCN++ \cite{Yu-CVPR-2020} & 4.2$\pm$3.4 & 5.8$\pm$4.9 & 5.0$\pm$2.9 \\ 
     & NAS-FAS \cite{Yu-PAMI-2020} & 4.2$\pm$5.3 & \underline{1.7$\pm$2.6} & 2.9$\pm$2.8 \\ 
     & PatchNet \cite{Wang-CVPR-2022} & \underline{2.5$\pm$3.81} & 3.33$\pm$3.73 & 2.9$\pm$3.0 \\ 
     & Watanabe \cite{Watanabe-APSIPA-2022} & 9.6$\pm$7.0 & 5.6$\pm$4.7 & 7.6$\pm$5.4 \\ 
     & Chen \cite{Chen-CoRR-2023} & --- & --- & 2.8$\pm$2.6 \\
     & TransFAS \cite{Wang-TBIOM-2022} & {\bf 2.1$\pm$2.2} & 3.8$\pm$3.5 & 2.9$\pm$2.4 \\ 
     & Li \cite{Li-NN-2024} & 2.9$\pm$2.9 & \underline{1.7$\pm$2.6} & \underline{2.3$\pm$2.2} \\
     & Proposed w/o PDA & 2.75$\pm$3.01 & {\bf 1.67$\pm$2.25} & {\bf 2.21$\pm$2.6} \\
     & Proposed w/ PDA & 2.75$\pm$2.41 & 2.33$\pm$2.39 & 2.54$\pm$2.38 \\
     \hline
    \end{tabular}
  }
\end{table}
\begin{table}[t]
  \centering
  \caption{Experimental results of each method for SiW (Unit: \%).}
  \label{tbl:results_SiW}
  \scalebox{0.8}[0.8]{
    \begin{tabular}{ccccc}
    \hline
    Prot. & Method & APCER$\downarrow$ & BPCER$\downarrow$ & ACER$\downarrow$ \\
    \hline
    \multirow{9}{*}{1} 
     & CDCN++ \cite{Yu-CVPR-2020} & \underline{0.07} & 0.17 & 0.12 \\ 
     & NAS-FAS \cite{Yu-PAMI-2020} & \underline{0.07} & 0.17 & 0.12 \\ 
     & PatchNet \cite{Wang-CVPR-2022} & {\bf 0.00} & {\bf 0.00} & {\bf 0.00} \\ 
     & Watanabe \cite{Watanabe-APSIPA-2022} & 0.11 & \underline{0.08} & 0.10 \\ 
     & TransFAS \cite{Wang-TBIOM-2022} & {\bf 0.00} & {\bf 0.00} & {\bf 0.00} \\
     & MFAE \cite{Zheng-IFS-2024} & {\bf 0.00} & {\bf 0.00} & {\bf 0.00} \\
     & Li \cite{Li-NN-2024} & {\bf 0.00} & 0.16 & \underline{0.08} \\
     & Proposed w/o PDA & 0.45 & 0.33 & 0.39 \\
     & Proposed w/ PDA & 0.1 & \underline{0.08} & 0.09 \\
     \hline
    \multirow{9}{*}{2} 
     & CDCN++ \cite{Yu-CVPR-2020} & {\bf 0.00$\pm$0.00} & 0.09$\pm$0.10 & 0.04$\pm$0.05 \\ 
     & NAS-FAS \cite{Yu-PAMI-2020} & {\bf 0.00$\pm$0.00} & 0.09$\pm$0.10 & 0.04$\pm$0.05 \\ 
     & PatchNet \cite{Wang-CVPR-2022} & {\bf 0.00$\pm$0.00} & {\bf 0.00$\pm$0.00} & {\bf 0.00$\pm$0.00} \\ 
     & Watanabe \cite{Watanabe-APSIPA-2022} & \underline{0.01$\pm$0.01} & \underline{0.01$\pm$0.01} & \underline{0.01$\pm$0.01} \\
     & TransFAS \cite{Wang-TBIOM-2022} & {\bf 0.00$\pm$0.00} & {\bf 0.00$\pm$0.00} & {\bf 0.00$\pm$0.00} \\
     & MFAE \cite{Zheng-IFS-2024} & {\bf 0.00$\pm$0.00} & {\bf 0.00$\pm$0.00} & {\bf 0.00$\pm$0.00} \\
     & Li \cite{Li-NN-2024} & {\bf 0.00$\pm$0.00} & 0.16$\pm$0.00 & 0.08$\pm$0.00 \\
     & Proposed w/o PDA & 0.05$\pm$0.07 & 0.05$\pm$0.05 & 0.05$\pm$0.06 \\
     & Proposed w/ PDA & 0.02$\pm$0.03 & 0.02$\pm$0.03 & 0.02$\pm$0.03 \\ \hline
    \multirow{9}{*}{3} 
     & CDCN++ \cite{Yu-CVPR-2020} & 1.97$\pm$0.33 & 1.77$\pm$0.10 & 1.90$\pm$0.15 \\ 
     & NAS-FAS \cite{Yu-PAMI-2020} & 1.58$\pm$0.23& 1.46$\pm$0.08 & 1.52$\pm$0.13 \\ 
     & PatchNet \cite{Wang-CVPR-2022} & 3.06$\pm$1.1 & 1.83$\pm$0.83 & 2.45$\pm$0.45 \\ 
     & Watanabe \cite{Watanabe-APSIPA-2022} & 3.07$\pm$2.75 & 3.07$\pm$2.75 & 3.07$\pm$2.75 \\
     & TransFAS \cite{Wang-TBIOM-2022} & 1.95$\pm$0.40 & 1.92$\pm$0.11 & 1.94$\pm$0.26 \\
     & MFAE \cite{Zheng-IFS-2024} & 2.57$\pm$1.83 & 1.92$\pm$1.06 & 2.42$\pm$1.45 \\
     & Li \cite{Li-NN-2024} & 2.13$\pm$1.22 & 2.25$\pm$1.06 & 2.19$\pm$1.14 \\
     & Proposed w/o PDA & \underline{1.34$\pm$0.58} & \underline{1.34$\pm$0.59} & \underline{1.34$\pm$0.58} \\
     & Proposed w/ PDA & {\bf 0.83$\pm$0.13} & {\bf 0.84$\pm$0.14} & {\bf 0.83$\pm$0.13} \\
     \hline
    \end{tabular}
  }
\end{table}
\subsection{Exp. (v): Comparison between Proposed and Conventional Methods}
\label{sec:comparison}

In this experiment, we compare the proposed method, whose effectiveness has been demonstrated through the above ablation study, with conventional methods for detecting face spoofing attacks.
We used the following conventional methods: CDCN++ \cite{Yu-CVPR-2020}, NAS-FAS \cite{Yu-PAMI-2020}, PatchNet \cite{Wang-CVPR-2022}, Watanabe \cite{Watanabe-APSIPA-2022}, Chen \cite{Chen-CoRR-2023}, TransFAS \cite{Wang-TBIOM-2022}, MFAE \cite{Zheng-IFS-2024} and Li \cite{Li-NN-2024}.
Since ACER of OULU-NPU Protocol 4 is lower when PDA is not introduced as shown in Table \ref{tbl:Ablation_4}, we evaluate the detection accuracy of the proposed method with and without PDA in this experiment.
Table \ref{tbl:results_OULU} shows the experimental results for OULU-NPU, where the highest accuracy is shown in bold, and the second highest accuracy is underlined.
In protocol 1, Proposed w/o PDA is less accurate than the conventional methods, while Proposed w/ PDA is comparable to the conventional methods.
For protocols 2 and 3, Proposed w/o PDA is comparable to the conventional methods, while Proposed w/ PDA is more accurate than the conventional methods.
Focusing on ACER, Proposed w/ PDA achieves ACER below 0.6\%, while the lowest ACER of the conventional methods is 0.9\%.
This indicates that the proposed method is robust against unknown papers, devices, and different cameras in detecting spoofing attacks.
In Protocol 4, Proposed w/o PDA has the lowest ACER, and Proposed w/ PDA has a lower ACER than the conventional methods except Li \cite{Li-NN-2024}.
Protocol 4 is the most difficult evaluation protocol in OULU-NPU, and the effectiveness of the proposed method is demonstrated in this protocol.
Table \ref{tbl:results_SiW} shows the experimental results for SiW, where the highest accuracy is shown in bold, and the second highest accuracy is underlined.
In Protocol 1 and 2, all the methods exhibit low values of the evaluation metrics, and therefore, can detect spoofing attacks with high accuracy.
In other words, Protocol 1 and 2 are easy evaluation protocols.
Protocol 3, on the other hand, evaluates the detection accuracy of unknown spoofing attacks and thus the value of the evaluation metrics becomes higher.
The lowest ACER of the conventional methods is 1.52$\pm$0.13\%.
ACER of Proposed w/o PDA is 1.34$\pm$0.58\%, which is lower than that of the conventional methods.
In particular, ACER of Proposed w/ PDA is 0.83$\pm$0.13\%, which is significantly lower than that of the conventional methods.
This indicates that the proposed method is more robust against unknown spoofing attacks than the conventional methods.
These results demonstrate the effectiveness of using the intermediate features of ViT and introducing data augmentation specialized for spoofing attack detection.


\section{Conclusion}

In this paper, we proposed a face spoofing attack detection method that uses features extracted from the intermediate layers of ViT. 
The proposed method utilizes the features extracted from the intermediate layers of ViT, and introduces loss functions and a score calculation method using the intermediate features.
The proposed method also employs data augmentation methods enhancing the performance of face spoofing attack detection: PDA and FAS-Aug.
We demonstrated the effectiveness of the proposed method through the ablation study of the proposed method and the comparison with conventional methods using OULU-NPU and SiW. 
It is expected that the intermediate features of ViT can be used to improve the accuracy of other spoofing attack detection methods, such as our method.



\section{Acknowledgment}

This work was supported in part by JSPS KAKENHI JP 23H00463 and 23K28085, and JST Moonshot R\&D Grant Number JPMJMS2215.


\end{document}